\documentclass[letterpaper]{article} 
\usepackage[preprint]{aaai2027}
\usepackage[hyphens]{url}  
\usepackage{graphicx} 
\usepackage{natbib}  
\usepackage{caption} 
\usepackage{algorithm}
\usepackage{algorithmic}
\usepackage{booktabs}
\usepackage{amssymb}
\ifdefined\pdfmapfile
\pdfmapfile{+newtx.map}
\pdfmapfile{+qtm-ts1.map}
\fi
\newcommand{\method}{PTQ4SNN}
\newcommand{\bridge}{Unified Scale Bridge}
\newcommand{\mpba}{Mixed-Precision Bit Allocation}

\newcommand{\accdrop}[2]{#1\,(#2)}

\title{\method: Membrane-Aware Post-Training Quantization for Spiking Neural Networks}
\author{
\textbf{Hui Xie}\textsuperscript{*}$^{1,2}$,
\textbf{Tong Shi}\textsuperscript{*}$^{1,2}$,
\textbf{Haotong Qin}$^{3}$,\\
\textbf{Aishan Liu}$^{1,4}$,
\textbf{Xiaode Liu}$^{5}$,
\textbf{Jinyang Guo}\textsuperscript{\dag}$^{1,2}$
}
\affiliations{
$^{1}$State Key Laboratory of Complex \& Critical Software Environment, Beihang University\\
$^{2}$School of Artificial Intelligence, Beihang University\\
$^{3}$Center for Project-Based Learning, D-ITET, ETH Zurich\\
$^{4}$School of Computer Science and Engineering, Beihang University\\
$^{5}$Intelligent Science \& Technology Academy of CASIC\\
\texttt{\small xiehui@buaa.edu.cn, shitong05@buaa.edu.cn, qinhaotong@gmail.com}\\
\texttt{\small liuaishan@buaa.edu.cn, lxde@pku.edu.cn, jinyangguo@buaa.edu.cn}\\
\textsuperscript{*}Equal contribution. \textsuperscript{\dag}Corresponding author.
}

\begin{document}

\maketitle

\begin{abstract}
Spiking neural networks (SNNs) enable sparse and event-driven computation, but their low-bit deployment remains incomplete because recurrent membrane states are commonly retained in floating point even after weight quantization. Quantizing these states is challenging because their distributions differ across channels and from the preceding weights, while small perturbations near the firing threshold may alter spike decisions and accumulate over time. We propose \method, a membrane-aware post-training quantization framework that jointly quantizes weights and recurrent membrane states using only a small calibration set. First, a channel-wise \bridge\ constrains the membrane scale as
\(s_{\mathrm{mem},c}=s_{w,c}2^{k_c}\), adapting to membrane distributions while enabling shift-compatible scale conversion. Second, \mpba\ assigns 2/4/8-bit precision to membrane channels according to firing activity and quantization sensitivity under an average-bit budget. The framework operates on reusable projection--LIF pairs and supports both convolutional SNNs and spike-driven Transformers without backbone retraining. Experiments on static and event-based classification and semantic segmentation show that \method\ effectively preserves model accuracy under W4 quantization and approximately 4-bit membrane precision.
Our code is publicly available at \url{https://github.com/dongfengwuyishi/ptq4snn}.
\end{abstract}

\section{Introduction}

Spiking neural networks (SNNs) have attracted increasing attention as an
energy-efficient alternative to conventional neural networks because they
communicate through sparse binary spikes and support event-driven computation
\cite{gerstner2002spiking,izhikevich2003simple,roy2019towards}.
Recent advances in residual SNNs and spike-driven Transformers, such as
Spikformer, QKFormer, Spike-Driven Transformer (SDT), and
Meta-SpikeFormer, have substantially improved performance on large-scale
vision tasks
\cite{fang2021deep,zhou2023spikformer,zhou2024qkformer,
yao2023spikedriven,yao2024spikedrivenv2}.
However, the increasing model size and feature resolution of these
architectures also introduce considerable parameter and neuronal-state
storage, limiting their deployment on resource-constrained platforms.

\begin{figure}[t]
  \centering
  \includegraphics[width=0.98\columnwidth]{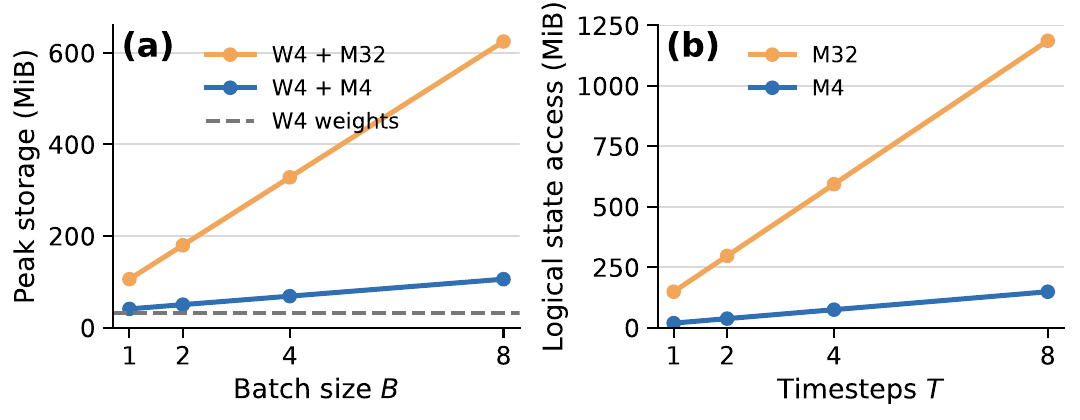}
  \caption{Membrane-state cost for SDT-8-768 at $224\times224$ under ideally
  packed W4 Conv/Linear weights. (a) Peak weight-plus-state storage versus
  batch size $B$. (b) Logical state read/write volume versus $T$ at $B=1$.
  M4 reduces the membrane contributions by $8\times$; actual traffic depends
  on caching and scheduling.}
  \label{fig:membrane_motivation}
\end{figure}

Among the recurrent states of an SNN, the membrane potential is particularly
important. At each timestep, an LIF neuron integrates the current synaptic
input with its previous state, generates a spike when the firing threshold is
crossed, and carries the updated membrane value to the next timestep
\cite{maass1997networks,gerstner2002spiking,neftci2019surrogate}.
Although spikes are binary, membrane states are commonly maintained in
floating point during inference \cite{yin2024mint,wei2024qsnn}.
Consequently, weight-only quantization leaves a substantial portion of the
state storage and recurrent data movement uncompressed, as illustrated in
Figure~\ref{fig:membrane_motivation}.

\begin{figure*}[t]
\centering
\includegraphics[width=0.98\textwidth]{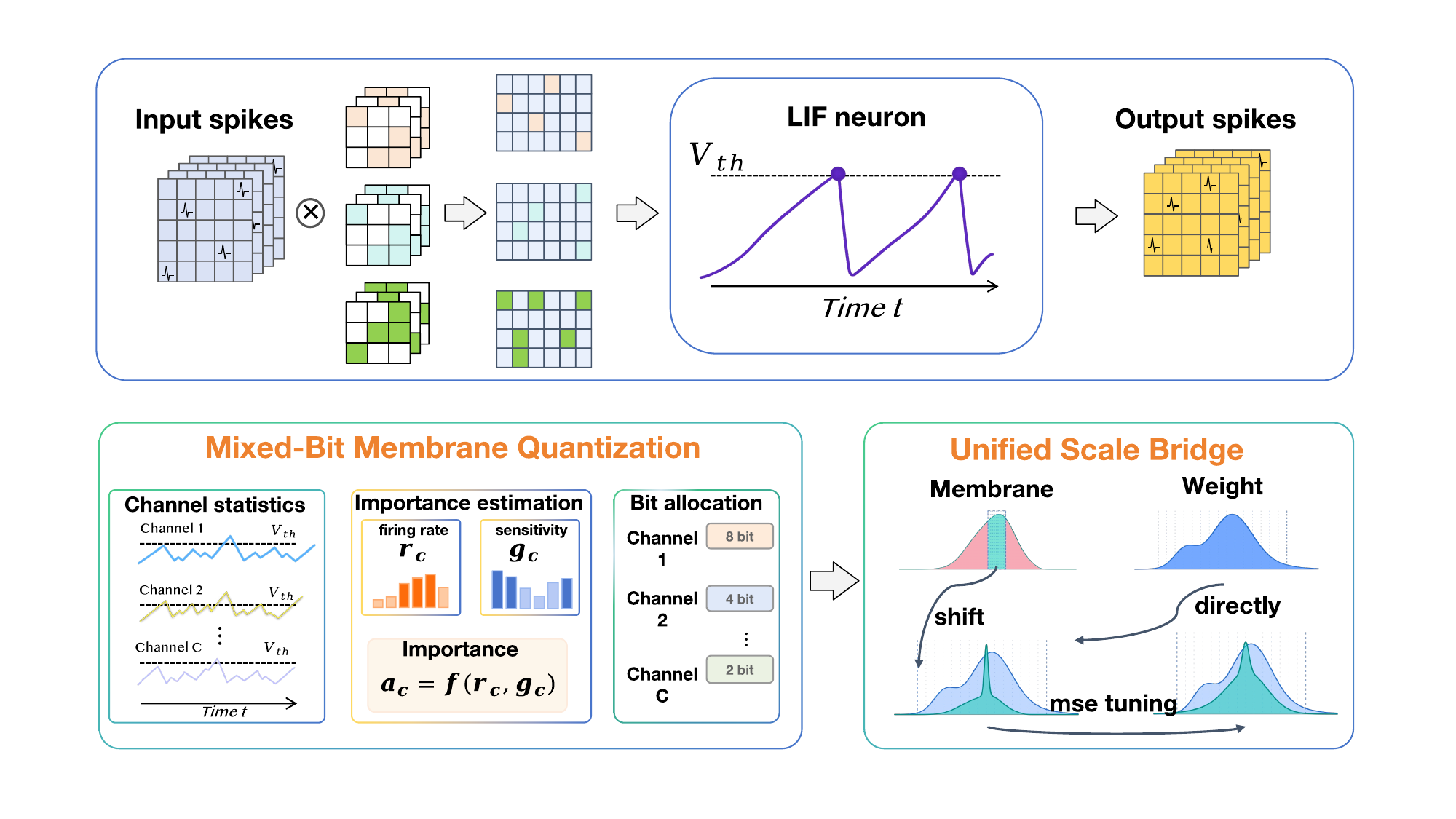}
\caption{Overview of \method. For each projection--LIF pair, channel-wise membrane statistics provide firing-rate and sensitivity estimates for mixed-bit allocation. The Unified Scale Bridge then couples each membrane scale to the preceding weight scale through shift-compatible conversion, conditioned on the assigned membrane bit width.}
\label{fig:framework_overview}
\end{figure*}

Membrane quantization presents challenges beyond ordinary activation
quantization. First, membrane distributions may differ substantially from
the distributions and scales of their preceding weights. Directly sharing
the weight scale can therefore cause severe clipping, whereas independently
quantizing the membrane introduces additional scale-conversion overhead.
Second, membrane errors are threshold-sensitive: a small perturbation can
change the current spike decision and subsequently alter the temporal state
trajectory. Finally, membrane channels exhibit different firing activities
and sensitivities, making a uniform bit width inefficient.

Existing SNN quantization methods mainly rely on quantization-aware training
or focus on weight quantization
\cite{yin2024mint,wei2024qsnn,qiu2025quantized,li2025neuronquant}.
FlowQ extends post-training quantization to membrane states using layer-wise
power-of-two scale coupling \cite{asim2026flowq}, but does not account for
channel-level scale and precision heterogeneity. Therefore, a unified PTQ
method is still needed to adapt both the scale and bit width of recurrent
membrane states while preserving hardware-friendly integer computation.

To address these issues, we propose \method, a mem\-brane-aware PTQ framework
for both conventional SNNs and spike-driven Transformers. We organize
quantizable projections and their following LIF neurons as reusable
projection--LIF pairs. For each pair, a channel-wise \bridge\ constrains the
membrane scale by
\(s_{\mathrm{mem},c}=s_{w,c}2^{k_c}\), allowing the quantization range to
adapt to the membrane distribution while retaining shift-compatible
conversion. We further introduce \mpba, which assigns 2/4/8-bit membrane
precision according to channel firing activity and quantization sensitivity
under an element-count-weighted average-bit budget.

Our main contributions are:

\begin{itemize}
    \item We introduce a channel-wise \bridge\ for membrane-aware PTQ,
    adapting to weight--membrane distribution mismatch while preserving
    power-of-two, shift-compatible scale conversion.

    \item We propose activity- and sen\-si\-tiv\-ity-aware Mixed-Pre\-ci\-sion Bit Allocation for recurrent
    membrane states, which assigns 2/4/8-bit precision at channel level
    under an element-count-weighted average-bit budget.

    \item We conduct extensive experiments on diverse SNN architectures and
    tasks, including conventional convolutional SNNs and spike-driven
    Transformers for static-image classification, event-based recognition,
    and semantic segmentation. The results demonstrate that \method\
    consistently preserves competitive accuracy under W4 quantization and
    an approximately 4-bit membrane budget, validating its effectiveness
    and generality across architectures and tasks.
\end{itemize}

\section{Related Work}

\textbf{SNN architectures.}
Conventional SNNs commonly adopt VGG- or residual-style backbones and use leaky integrate-and-fire (LIF) neurons for temporal computation \cite{sengupta2019going,fang2021deep}.
Spiking Transformers extend SNNs to token-based visual modeling.
Spikformer introduces spike-based self-attention, QKFormer uses spike-based query--key interactions, and the Spike-Driven Transformer (SDT) series develops scalable spike-driven attention and meta-architectures \cite{zhou2023spikformer,zhou2024qkformer,yao2023spikedriven,yao2024spikedrivenv2}.
Despite their different operators, both conventional SNNs and SDTs repeatedly update recurrent membrane states.

\textbf{Quantization-aware training for SNNs.}
MINT jointly quantizes weights and membrane potentials with a shared scale for multiplier-less inference \cite{yin2024mint}.
SQUAT introduces uniform and threshold-centered quantizers for stateful neurons during training \cite{venkatesh2024squat}.
Q-SNNs combines binary weights and low-bit membrane potentials with weight--spike dual regulation \cite{wei2024qsnn}.
QSD-Transformer applies low-bit weight training to spike-driven Transformers and introduces an information-enhanced LIF neuron and fine-grained distillation \cite{qiu2025quantized}.
HardF-SNN develops proportional shared-scale quantization and integer-only batch normalization for hardware-friendly inference \cite{liu2026hardfsnn}.
These methods recover low-bit accuracy by adapting model parameters during training.
However, they require additional training or retraining, often with the original training data.

\textbf{Post-training quantization for SNNs.}
NeuronQuant introduces neuron-wise calibration for pretrained SNNs but does not quantize membrane potentials \cite{li2025neuronquant}.
SNNQ improves ultra-low-bit weight PTQ through block-wise weight rounding and test-time neuron pruning \cite{zhang2026snnq}.
FlowQ jointly quantizes weights and membrane potentials using MSE-calibrated layer-wise scales constrained by a power-of-two ratio \cite{asim2026flowq}.
NeuronQuant and SNNQ do not target recurrent membrane-state quantization.
FlowQ compresses membrane states, but uses layer-wise scale coupling and one membrane bit width per setting.
Existing SNN PTQ therefore does not fully address channel-level membrane heterogeneity across both conventional SNNs and SDTs.

\textbf{Mixed-precision quantization for SNNs.}
Hessian-aware quantization assigns layer-wise bit widths according to loss curvature and fine-tunes the quantized model \cite{lui2021hessian}.
Q-SpiNN searches quantization schemes, precision levels, and rounding strategies for different SNN parameter types under both post-training and in-training settings \cite{putra2021qspinn}.
These methods operate at layer or parameter-type granularity and are evaluated on relatively small conventional SNNs.
They do not perform calibration-only, activity- and sensitivity-aware bit allocation for recurrent membrane channels.

\begin{figure}[t]
\centering
\includegraphics[width=0.78\columnwidth]{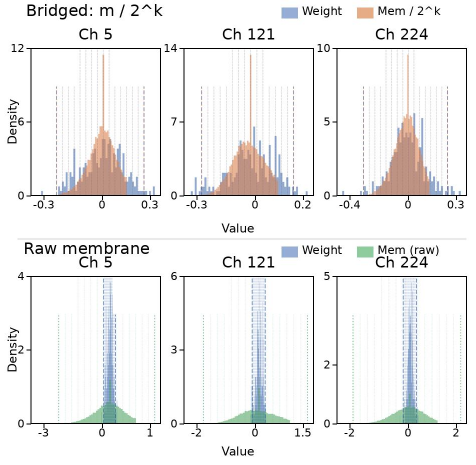}
\caption{Weight and membrane distributions across three representative channels of a paired projection-LIF unit. Raw membrane states often occupy a wider range than weights, while the Unified Scale Bridge aligns each membrane grid with the weight-domain scale through \(2^k\) without using a general multiplier.}
\label{fig:distribution_analysis}
\end{figure}

\section{Preliminaries}

\paragraph{Spiking neuron model.}
Spiking neurons process information over discrete timesteps through membrane integration, spike generation, and reset
\cite{gerstner2002spiking,izhikevich2003simple,neftci2019surrogate,wu2018stbp}.
We describe these dynamics using the widely adopted leaky integrate-and-fire (LIF) model.
For layer $\ell$ at timestep $t$, the neuronal update is written as
\begin{eqnarray}
\mathbf{I}^{\ell}[t]
&=& f\!\left(\mathbf{W}^{\ell},\mathbf{S}^{\ell-1}[t]\right), \\
\widetilde{\mathbf{V}}^{\ell}[t]
&=& \tau\mathbf{V}^{\ell}[t-1]+\mathbf{I}^{\ell}[t], \\
\mathbf{S}^{\ell}[t]
&=& H\!\left(\widetilde{\mathbf{V}}^{\ell}[t]-\theta^{\ell}\right), \\
\mathbf{V}^{\ell}[t]
&=& \mathcal{R}\!\left(
\widetilde{\mathbf{V}}^{\ell}[t],
\mathbf{S}^{\ell}[t]
\right),
\end{eqnarray}
where $\mathbf{W}^{\ell}$ denotes the synaptic weight,
$f(\cdot)$ denotes a convolutional or linear transformation,
$\tau\in(0,1]$ is the membrane leak factor, and
$\theta^{\ell}$ is the firing threshold.
$H(\cdot)$ is the Heaviside step function, so the output spike
$\mathbf{S}^{\ell}[t]$ is binary.
The reset function $\mathcal{R}(\cdot)$ is commonly implemented as
\begin{equation}
\mathcal{R}(\widetilde{\mathbf{V}},\mathbf{S})=
\left\{
\begin{array}{ll}
\widetilde{\mathbf{V}}(1-\mathbf{S}),
& \mathrm{hard~reset}, \\
\widetilde{\mathbf{V}}-\theta\mathbf{S},
& \mathrm{soft~reset}.
\end{array}
\right.
\end{equation}
Here, $\widetilde{\mathbf{V}}^{\ell}[t]$ is the membrane potential
before spike generation, while $\mathbf{V}^{\ell}[t]$ is the
post-reset state carried to the next timestep.

\paragraph{Conventional SNN.}
Conventional SNNs commonly adopt VGG-style or residual convolutional
backbones \cite{sengupta2019going,fang2021deep}.
An input is represented over $T$ timesteps, and each layer applies a
convolutional or linear transformation followed by a spiking neuron.
For static images, the input may be repeated or encoded into a temporal
sequence; for event-based data, events are usually aggregated into
discrete temporal bins.
The network prediction is obtained by accumulating or averaging the
output features or logits over time.

\paragraph{Spike-driven transformer.}

Recent studies extend SNNs to Transformer architectures by introducing
spike-driven computation for token interaction and feature extraction
\cite{yao2023spikedriven,yao2024spikedrivenv2,yao2025scaling}.
Unlike conventional artificial Transformers, spike-driven transformers
replace dense floating-point activation processing with sparse spike
communication while maintaining the hierarchical structure of
Transformer blocks.

A typical spike-driven transformer consists of projection layers,
spike-driven attention modules, MLP blocks, and LIF neurons.
Given an input feature sequence $\mathbf{X}$, linear projections are
first applied to generate intermediate features, which are then
processed by spiking neurons:
\begin{equation}
\mathbf{S}_t=\mathcal{SN}(\mathbf{W}\mathbf{X}_t),
\end{equation}
where $\mathcal{SN}(\cdot)$ denotes an LIF neuron.
The membrane potential generated by the LIF neuron is retained across
timesteps and participates in subsequent spike generation.

Although spike-driven transformers employ different attention and
feature mixing designs, they share the same temporal computation
principle with conventional SNNs: intermediate features are integrated
into recurrent membrane states through LIF neurons.
Therefore, membrane states are essential components for both
convolutional SNNs and spike-driven transformers.

\paragraph{Post-training quantization.}
Post-training quantization (PTQ) converts a pretrained floating-point
model to low precision without retraining its network parameters
\cite{jacob2018quantization,nagel2020adaround,li2021brecq}.
For a real-valued tensor $\mathbf{X}$, symmetric uniform quantization
with bit width $b$ is defined as
\begin{eqnarray}
\mathbf{X}_{\mathrm{int}}
&=&
\mathrm{clip}\!\left(
\left\lfloor \mathbf{X}/s_{X}\right\rceil,
q_{\min},q_{\max}
\right), \\
\widehat{\mathbf{X}}
&=&
s_{X}\mathbf{X}_{\mathrm{int}},
\end{eqnarray}
where $s_{X}$ is the quantization scale,
$q_{\min}=-2^{b-1}$, and $q_{\max}=2^{b-1}-1$ for signed quantization.
The scale and clipping range are estimated from the pretrained
parameters or a small calibration set and are fixed during inference.
We use W$b_w$/M$b_m$ to denote $b_w$-bit weights and $b_m$-bit
membrane states, respectively.

\section{Method}

\subsection{Problem Definition}

We organize each quantizable projection and its subsequent LIF neuron
as a projection--LIF pair. The projection may be a convolution, a
linear layer, a Q/K/V projection, an MLP projection, or a
residual-branch projection. Given a pretrained SNN and a small
calibration set, \method\ keeps the backbone parameters fixed and
determines only the quantization variables used during inference.

\subsection{Framework Overview}

Figure~\ref{fig:framework_overview} gives an overview of \method, which proceeds in three stages. First, it prepares a quantized model by folding normalization where applicable and replacing paired Conv/Linear-LIF modules with quantized counterparts. Second, it performs calibration forward passes to collect weight scales, membrane ranges, firing-rate statistics, and quantization-sensitivity statistics. Third, it first assigns channel-wise membrane precision under an average-bit budget and then constructs the Unified Scale Bridge conditioned on the assigned bit width of each channel.

\begin{algorithm}[t]
\caption{\method\ PTQ Pipeline}
\label{alg:ptq4spike}
\begin{algorithmic}[1]
\REQUIRE pretrained spiking model \(f\), calibration set \(\mathcal{D}_c\), weight bit \(b_w\), membrane budget \(b_m\)
\STATE Build the quantized model \(f_q\) from \(f\)
\STATE Collect weight scales, membrane ranges, firing-rate statistics, and quantization-sensitivity statistics on \(\mathcal{D}_c\)
\IF{mixed-precision membrane allocation is enabled}
\STATE Compute channel scores from firing rates and quantization sensitivity
\STATE Assign \(b_c\in\{2,4,8\}\) under the average membrane-bit budget
\ELSE
\STATE Set \(b_c=b_m\) for all membrane channels
\ENDIF
\STATE For each paired projection--LIF unit, choose \(k_c\) conditioned on \(b_c\) and set \(s_{\mathrm{mem},c}=s_{w,c}2^{k_c}\)
\STATE Freeze quantization parameters and evaluate \(f_q\)
\end{algorithmic}
\end{algorithm}

\subsection{Unified Scale Bridge}

The Unified Scale Bridge creates a hardware-friendly relation between the preceding weight scale and the membrane scale:
\begin{equation}
s_{\mathrm{mem},c}=s_{w,c}2^{k_c}.
\end{equation}
Compared with direct scale reuse \(s_{\mathrm{mem}}=s_w\), this relation lets the membrane quantization step expand or shrink according to calibration statistics. Compared with an independent membrane scale, the relative conversion is a power of two and can be implemented as a shift. Since the representable integer range depends on the assigned membrane bit width \(b_c\), the bridge exponent \(k_c\) is calibrated after membrane-bit allocation.
Figure~\ref{fig:distribution_analysis} illustrates the weight-membrane distribution mismatch that motivates this bridge.

\begin{figure}[t]
\centering
\includegraphics[width=0.98\columnwidth]{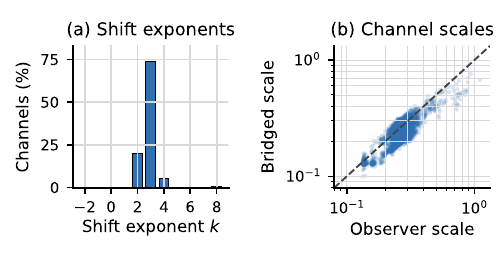}
\caption{Membrane scale analysis. (a) Distribution of the channel-wise integer shift exponents \(k_c\) selected by the Unified Scale Bridge after membrane-bit allocation. (b) Bridged membrane scales versus independently calibrated observer scales; the dashed line marks exact agreement.}
\label{fig:bridge_analysis}
\end{figure}

Let integer weight and membrane values be \(\bar{W}\) and \(\bar{V}\). Since
\begin{equation}
V=s_{\mathrm{mem}}\bar{V}=s_w2^k\bar{V},
\end{equation}
the membrane term can be mapped into the weight-scale domain by shifting \(\bar{V}\). If leakage is represented with a fixed-point or power-of-two approximation, the pre-fire update can be expressed as
\begin{equation}
\bar{V}^{(w)}_{\mathrm{pre},t}
=\bar{W}x_t+\tau 2^k\bar{V}^{(\mathrm{mem})}_{t-1}.
\end{equation}
In SDT experiments, the bridge is searched under the assigned channel-wise membrane bit width using a combined quantization error:
\begin{equation}
\min_{s_{w,c},k_c}
\mathcal{E}_{w}(W_c;s_{w,c})
+
\lambda_{\mathrm{mem}}
\mathcal{E}_{\mathrm{mem}}
\!\left(
V_c;
s_{w,c}2^{k_c},
b_c
\right).
\end{equation}
Here, \(b_c\) determines the signed integer range used to quantize channel \(c\), and therefore directly affects the optimal shift exponent \(k_c\). In conventional SNN experiments, the same principle is implemented as interval-calibrated membrane scaling: a power-of-two factor shifts membrane values into a better shared-scale integer interval under the assigned bit width, while the threshold comparison is shifted consistently.
Figure~\ref{fig:bridge_analysis} characterizes the calibrated bridge. Panel (a) reports the channel-wise integer shift exponents selected by the bridge, while panel (b) compares the resulting membrane scales with independently calibrated observer scales. Their concentration near the identity line indicates that shift-compatible coupling closely tracks the observer-calibrated channel scales.

\subsection{Mixed-Precision Membrane Bit Allocation}

SNN membrane channels are heterogeneous across architectures: some channels fire frequently while others are inactive, and some channels are more sensitive to quantization perturbation. In SDTs this heterogeneity appears in patch/stem projections, Q/K/V projections, MLP projections, and SepConv paths; in conventional SNNs it appears in Conv-LIF and residual branches. \mpba\ assigns membrane bit-widths per channel under a weighted average budget, rather than forcing all channels to uniform M4.

\begin{figure}[t]
\centering
\includegraphics[width=0.96\columnwidth]{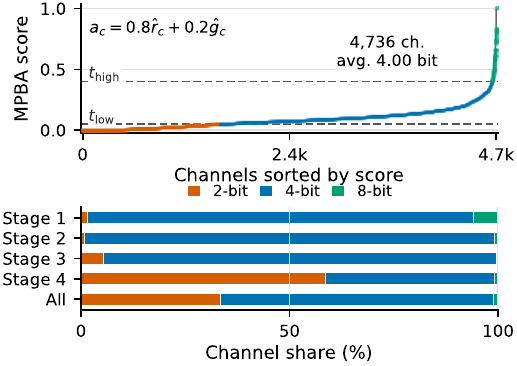}
\caption{Mixed-precision membrane allocation on SEW-ResNet18/CIFAR-10. The upper panel ranks real LIF channels by \(a_c=0.8\hat r_c+0.2\hat g_c\) and applies budgeted thresholds; the lower panel reports stage-wise 2/4/8-bit composition under a 4-bit average membrane budget.}
\label{fig:mpba_diagram}
\end{figure}

For channel \(c\), let \(S_c(x,t,u)\) denote the binary spike at timestep
\(t\) and spatial or token position \(u\). Its firing rate is
\begin{equation}
r_c=\frac{1}{|\mathcal{D}_c|T|\Omega_c|}
\sum_{x\in\mathcal{D}_c}\sum_{t=1}^{T}\sum_{u\in\Omega_c}S_c(x,t,u).
\end{equation}
A common 4-bit reference pass yields spike output \(S^q_{d,c}\), while a
quantization-disabled pass yields \(S^{\mathrm{FP}}_{d,c}\). For calibration
minibatch \(d\in\mathcal{B}_c\), we set \(\widetilde y_d=\arg\max f(x_d)\) and
\(\mathcal{L}_d=\mathrm{CE}(f_q(x_d),\widetilde y_d)\). With
\(\Delta S_{d,c}=S^q_{d,c}-S^{\mathrm{FP}}_{d,c}\), we define
\begin{eqnarray}
z_{d,c}&=&\left\langle\nabla_{S^q_{d,c}}\mathcal{L}_d,
\Delta S_{d,c}\right\rangle,\nonumber\\
g_c&=&\frac{1}{|\mathcal{B}_c|}\sum_{d\in\mathcal{B}_c}
\left|z_{d,c}+\frac{1}{2}z_{d,c}^{2}\right|.
\end{eqnarray}
This reference is fixed before bit allocation and bridge calibration. Across
eligible channels, min--max normalization and the combined score are
\begin{eqnarray}
\hat r_c&=&\frac{r_c-r_{\min}}{r_{\max}-r_{\min}+\epsilon},\nonumber\\
\hat g_c&=&\frac{g_c-g_{\min}}{g_{\max}-g_{\min}+\epsilon},\nonumber\\
a_c&=&\beta\hat r_c+\gamma\hat g_c,\qquad \beta+\gamma=1.
\end{eqnarray}
where the extrema are computed over all eligible channels.
Channels are assigned 2, 4, or 8 membrane bits by score quantiles. On a small held-out calibration subset, we grid-search the 8/4-bit boundary (the sparse-protection percentile) and the activity--sensitivity weight \(\beta\); the remaining 4/2-bit boundary is determined by binary search so that the element-count-weighted average membrane precision satisfies
\begin{equation}
\frac{\sum_c b_c N_c}{\sum_c N_c}\approx b_m,
\end{equation}
where \(N_c\) is the number of membrane values represented by channel \(c\). This restricted search is inexpensive. The first spiking layer, classifier-adjacent layers, and a small set of highly important channels can be protected with higher precision for stability. Although individual membrane channels may use 2, 4, or 8 bits, we report this operating point as M4 throughout the main experiments because its element-count-weighted average membrane precision is close to 4 bits. After \(b_c\) is assigned, it is passed to the Unified Scale Bridge, which searches the corresponding \(k_c\) under the channel-specific integer range.
Figure~\ref{fig:mpba_diagram} visualizes this score-ranking and budgeted thresholding procedure on a calibrated SEW-ResNet18 run.

The allocation is intentionally tied to LIF channels rather than to whole blocks. Whole-block precision would assign the same membrane bit-width to active and inactive channels inside a projection, which is inefficient for SDT attention projections and conventional residual branches. Channel-wise MPBA keeps the average precision fixed while moving bits toward channels that either fire often or have higher quantization sensitivity.

\begin{table*}[t]
\centering
\small
\setlength{\tabcolsep}{8.5pt}
\begin{tabular}{llccc}
\toprule
\textbf{Backbone} & \textbf{Method} &
\shortstack{\textbf{M4}\\\textbf{scheme}} &
\textbf{W4/M32} & \textbf{W4/M4} \\
\midrule
SDT-8-768 ($T=4$) & Full precision & -- &
\multicolumn{2}{c}{\accdrop{75.90}{Base}} \\
\cmidrule(lr){2-5}
& BRECQ~\cite{li2021brecq} & Reuse &
\accdrop{74.12}{-1.78} & \accdrop{4.84}{-71.06} \\
& GPTQ~\cite{frantar2023gptq} & Reuse &
\accdrop{74.11}{-1.79} & \accdrop{5.34}{-70.56} \\
& FlowQ~\cite{asim2026flowq} & LW-PoT &
-- & \accdrop{63.74}{-12.16} \\
& \textbf{\method} & CW-PoT + MPBA &
-- & \textbf{\accdrop{75.16}{-0.74}} \\
\midrule
Meta-SpikeFormer ($T=4$) & Full precision & -- &
\multicolumn{2}{c}{\accdrop{78.88}{Base}} \\
\cmidrule(lr){2-5}
& BRECQ~\cite{li2021brecq} & Reuse &
\accdrop{74.76}{-4.12} & \accdrop{68.66}{-10.22} \\
& GPTQ~\cite{frantar2023gptq} & Reuse &
\accdrop{77.66}{-1.22} & \accdrop{75.31}{-3.57} \\
& FlowQ~\cite{asim2026flowq} & LW-PoT &
-- & \accdrop{63.54}{-15.34} \\
& \textbf{\method} & CW-PoT + MPBA &
-- & \textbf{\accdrop{78.50}{-0.38}} \\
\midrule
SEW-ResNet18 ($T=4$) & Full precision & -- &
\multicolumn{2}{c}{\accdrop{62.69}{Base}} \\
\cmidrule(lr){2-5}
& BRECQ~\cite{li2021brecq} & Reuse &
\accdrop{60.37}{-2.32} & \accdrop{58.28}{-4.41} \\
& GPTQ~\cite{frantar2023gptq} & Reuse &
\accdrop{61.33}{-1.36} & \accdrop{56.85}{-5.84} \\
& FlowQ~\cite{asim2026flowq} & LW-PoT &
-- & \accdrop{58.30}{-4.39} \\
& \textbf{\method} & CW-PoT + MPBA &
-- & \textbf{\accdrop{59.62}{-3.07}} \\
\bottomrule
\end{tabular}
\normalsize
\caption{ImageNet-1K classification results. Each accuracy is followed by its change from the corresponding floating-point checkpoint. \textit{Reuse} denotes direct reuse of the preceding weight scale for membrane quantization, \(s_{\mathrm{mem},c}=s_{w,c}\). LW-PoT and CW-PoT denote layer-wise and channel-wise power-of-two weight--membrane scale coupling, respectively. For \method, W4/M4 denotes an average 4-bit membrane budget, implemented with channel-wise 2/4/8-bit allocation where applicable. FlowQ results are obtained from our reimplementation on the same checkpoints.}
\label{tab:imagenet_main}
\end{table*}

\section{Experiments}
\label{sec:experiments}

We evaluate \method\ on large-scale static-image classification,
event-based classification, and semantic segmentation. All experiments
start from pretrained checkpoints and use only calibration
samples without updating backbone parameters. Generic W4/M32 PTQ
baselines retain floating-point membrane states; their controlled W4/M4
extensions directly reuse the preceding weight scale for membrane
quantization. FlowQ is reimplemented with layer-wise power-of-two scale
coupling, whereas \method\ uses channel-wise coupling and is reported
under a W4/M4 budget. For \method, M4 denotes an element-count-weighted
average membrane precision of approximately 4 bits, with 2/4/8-bit
channel-wise allocation where applicable. Further details on datasets,
model configurations, calibration, and baseline protocols are available
in the supplementary document.

\subsection{Main Results}
\label{sec:main_results}

\paragraph{ImageNet-1K classification.}
Table~\ref{tab:imagenet_main} compares the methods on three ImageNet-1K
backbones. Existing W4/M32 baselines compress the weights but retain
full-precision recurrent states. By additionally quantizing the membrane
states, \method\ reaches 75.16\% on SDT-8-768 and 78.50\% on
Meta-SpikeFormer, corresponding to only 0.74 and 0.38 percentage-point
drops from their floating-point checkpoints, respectively. The
SEW-ResNet18 result further evaluates the method on a conventional
convolutional SNN. The W4/M4 entries compare direct weight-scale reuse,
FlowQ's layer-wise power-of-two coupling, and our channel-wise method
under the same average 4-bit membrane budget. On this backbone, the
completed BRECQ reuse baseline reaches 58.28\% at W4/M4.

\paragraph{Event-based classification.}
CIFAR10-DVS contains temporally varying event streams and is therefore a
direct test of recurrent membrane quantization. As shown in
Table~\ref{tab:dvs_main}, \method\ obtains 70.80\% at W4/M4, limiting
the drop from the floating-point checkpoint to 1.00 percentage point.
The W4/M4 column compares all membrane-quantized methods under the same
average 4-bit membrane budget.

\begin{table}[t]
\centering
\small
\setlength{\tabcolsep}{3.0pt}
\begin{tabular*}{\columnwidth}{@{\extracolsep{\fill}}lcc@{}}
\toprule
\textbf{Method} & \textbf{W4/M32} & \textbf{W4/M4} \\
\midrule
Full precision & \multicolumn{2}{c}{\accdrop{71.80}{Base}} \\
\midrule
BRECQ~\cite{li2021brecq} &
\accdrop{69.50}{-2.30} & \accdrop{61.70}{-10.10} \\
GPTQ~\cite{frantar2023gptq} &
\accdrop{70.20}{-1.60} & \accdrop{61.20}{-10.60} \\
FlowQ~\cite{asim2026flowq} &
-- & \accdrop{67.30}{-4.50} \\
\textbf{\method} &
-- & \textbf{\accdrop{70.80}{-1.00}} \\
\bottomrule
\end{tabular*}
\normalsize
\caption{CIFAR10-DVS classification with SDT-2-256 at $T=10$. Each entry reports Acc@1 (\%) and its change from the floating-point checkpoint. Generic W4/M4 baselines directly reuse the preceding weight scale for membrane quantization; FlowQ uses layer-wise power-of-two coupling, while \method\ uses channel-wise coupling and an average 4-bit membrane budget.}
\label{tab:dvs_main}
\end{table}

\begin{table}[t]
\centering
\small
\setlength{\tabcolsep}{0pt}
\begin{tabular*}{\columnwidth}{@{\extracolsep{\fill}}lccccc@{}}
\toprule
\textbf{Method} & \textbf{W/M} & \textbf{aAcc} &
\textbf{mIoU} & \textbf{mAcc} & $\Delta$ \\
\midrule
Full precision & 32/32 & 90.64 & 60.44 & 73.41 & 0.00 \\
\midrule
BRECQ~\cite{li2021brecq}
& 4/32 & 89.59 & 55.80 & 67.63 & -4.64 \\
& 4/4 & 82.73 & 33.92 & 44.47 & -26.52 \\
\cmidrule(lr){2-6}
GPTQ~\cite{frantar2023gptq}
& 4/32 & 90.33 & 58.41 & 70.12 & -2.03 \\
& 4/4 & 90.28 & 59.63 & 72.33 & -0.81 \\
\cmidrule(lr){2-6}
FlowQ~\cite{asim2026flowq} & 4/4 & 87.14 & 49.41 & 60.95 & -11.03 \\
\midrule
\textbf{\method}
& 4/4 & \textbf{90.54} & \textbf{59.77} &
\textbf{72.49} & \textbf{-0.67} \\
\bottomrule
\end{tabular*}
\normalsize
\caption{Semantic segmentation on Pascal VOC2012. $\Delta$ denotes the mIoU change from the floating-point checkpoint. Generic W4/M4 baselines directly reuse the preceding weight scale for membrane quantization. For \method, M4 denotes an average 4-bit membrane budget with channel-wise allocation.}
\label{tab:seg_main}
\end{table}

\paragraph{Semantic segmentation.}
We further apply the calibration-only framework to an SDT-based FPN on
Pascal VOC2012. Table~\ref{tab:seg_main} reports the dense-prediction
results. \method\ achieves 59.77 mIoU at W4/M4, only 0.67 point below
the floating-point model, whereas the layer-wise FlowQ baseline reaches
49.41 mIoU, an 11.03-point drop. This indicates that membrane-aware PTQ
is not tied to a classification head and transfers to dense prediction
without task-specific retraining. Qualitative results are provided in
the supplementary document.

\paragraph{Packed-state cost on a hardware model.}
We additionally estimate the resident-state cost of the calibrated
SEW-ResNet18/CIFAR-10 run. Following FlowQ's hardware-cost
methodology~\cite{asim2026flowq}, we evaluate an 8$\times$16-PE hardware
model with 32-bit SRAM words and use 32-nm, 400-MHz CACTI estimates. We
report normalized state-SRAM read/write energy. Membrane packets are
channel-major and padded to 32-bit words. The stem is deliberately retained
at 16 bits; all other 4,736 LIF channels are packed at their assigned
precision. Thus, unlike the nominal M4 label,
Table~\ref{tab:packed_state_cost} reports both the element-count-weighted
non-stem precision and the true average over all resident states. For MPBA,
each non-stem channel stores a 2-bit precision tag and an 8-bit bridge shift.
The payloads require no padding because every channel extent is a multiple of
a 32-bit word; the 5.78-KiB metadata is included in the packed total and
energy proxy.

\begin{table}[t]
\centering
\small
\setlength{\tabcolsep}{1.0pt}
\begin{tabular*}{\columnwidth}{@{\extracolsep{\fill}}lccccc@{}}
\toprule
\textbf{Mode} & \textbf{2/4/8} & $b_{\rm nonstem}$ & $b_{\rm all}$ &
\textbf{State (MiB)} & $E_{\rm R/W}$ \\
\midrule
M32 & 0/0/100 & 32.000 & 30.293 & 2.219 / 2.219 & 1.000 \\
Uniform M4 & 0/100/0 & 4.000 & 5.280 & 0.387 / 0.387 & 0.174 \\
MPBA & 6.20/90.65/3.15 & 4.002 & 5.282 &
0.387 / 0.392 & 0.177 \\
\bottomrule
\end{tabular*}
\normalsize
\caption{Packed resident membrane state on the 8$\times$16-PE hardware
model for SEW-ResNet18/CIFAR-10 ($B=1$). ``Logical'' excludes packing
metadata; ``packed'' includes 32-bit alignment, 2-bit precision tags, and
8-bit shifts. The 2/4/8 split and $b_{\rm nonstem}$ exclude the protected
16-bit stem, whereas $b_{\rm all}$ is the true average over all 614,400
state values. Energy is normalized state-SRAM read/write energy.}
\label{tab:packed_state_cost}
\end{table}

\subsection{Ablation Studies}
\label{sec:ablation}

\paragraph{Unified Scale Bridge.}
We first isolate membrane-scale construction on SEW-ResNet18/CIFAR-100
under W4/M4. As shown in Table~\ref{tab:ablation_scale}, direct weight-scale
reuse suffers severe clipping. An independent observer largely avoids
saturation, i.e., clipping at the quantization bounds, but requires an
unconstrained scale conversion. Here, an observer denotes a separately
calibrated membrane scale. The proposed
\bridge\ improves Acc@1 by 1.66 points over reuse and by 0.17 points over
the independent observer, while retaining shift-compatible conversion.

\begin{table}[t]
\centering
\small
\setlength{\tabcolsep}{7pt}
\begin{tabular}{lccc}
\toprule
\textbf{Scale construction} &
\textbf{Acc@1} &
\textbf{Drop} &
\textbf{Saturation (\%)} \\
\midrule
Reuse & 72.53 & 2.74 & 56.50 \\
Independent observer & 74.02 & 1.25 & 3.58 \\
\textbf{Unified Bridge} & \textbf{74.19} & \textbf{1.08} & 5.31 \\
\bottomrule
\end{tabular}
\normalsize
\caption{Membrane-scale construction under W4/M4. Drop is measured
from the 75.27\% floating-point checkpoint.}
\label{tab:ablation_scale}
\end{table}

\paragraph{Mixed-Precision Membrane Bit Allocation.}
We next evaluate the independent contribution of MPBA on
SEW-ResNet18. Table~\ref{tab:ablation_mpba_component} keeps W4, $T=4$,
and an average M4 membrane budget fixed. MPBA improves Acc@1 by 0.440
points on CIFAR-10 and 0.522 points on ImageNet-1K, showing that
channel-wise allocation uses the same average bit budget more
effectively than uniform precision.

\begin{table}[t]
\centering
\small
\setlength{\tabcolsep}{6pt}
\begin{tabular*}{\columnwidth}{@{\extracolsep{\fill}}lcc@{}}
\toprule
\textbf{Dataset} & \textbf{Acc@1 with MPBA} & \textbf{Gain} \\
\midrule
CIFAR-10 & \textbf{93.36} & +0.440 \\
ImageNet-1K & \textbf{59.62} & +0.522 \\
\bottomrule
\end{tabular*}
\normalsize
\caption{Independent MPBA effect on SEW-ResNet18 under W4/M4 and
$T=4$. Gain is measured relative to the corresponding run without
MPBA (percentage points).}
\label{tab:ablation_mpba_component}
\end{table}

\begin{figure}[ht]
\centering
\includegraphics[width=0.60\columnwidth]{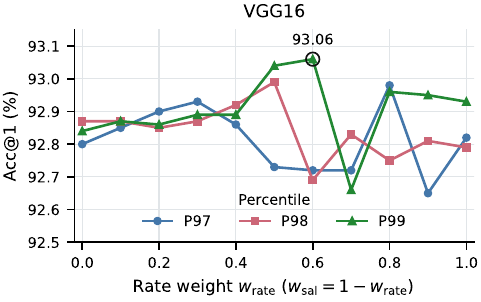}
\caption{MPBA hyperparameter sensitivity of VGG16 on CIFAR-10. Curves sweep
the firing-rate weight while the sensitivity weight is its complement;
P97--P99 denote the percentile used for sparse high-precision protection.}
\label{fig:mpba_sweep_vgg}
\end{figure}

\paragraph{MPBA hyperparameters.}
Scanning $\beta$ with $\gamma=1-\beta$ and the sparse-protection percentile
at W4/M4 and $T=4$ on VGG16 selects P99 with $\beta=0.6$ (93.06\%);
the scan in Figure~\ref{fig:mpba_sweep_vgg} confirms that combining activity
and sensitivity is preferable to either signal alone.

\section{Conclusion}

We presented \method, a membrane-aware PTQ framework that quantizes
weights and recurrent membrane states without full-network retraining.
Its channel-wise \bridge\ provides shift-compatible scale adaptation,
while \mpba\ allocates membrane precision according to channel activity
and sensitivity. Results across convolutional SNNs, spike-driven
Transformers, event-based recognition, and semantic segmentation support
membrane states as first-class PTQ targets.

\begingroup
\small
\bibliography{aaai2027}
\endgroup

\clearpage
\appendix


\section{Experimental Details}
\label{app:setup}

\subsection{Datasets and Architectures}

We evaluate static-image classification on ImageNet-1K
\cite{deng2009imagenet}, event-based classification on CIFAR10-DVS
\cite{li2017cifar10dvs}, and semantic segmentation on Pascal VOC2012
\cite{everingham2010pascal}.
Supplementary classification experiments are conducted on CIFAR-10 and
CIFAR-100 \cite{krizhevsky2009learning}.

SDT-2-256 is used for CIFAR-10, CIFAR-100, and CIFAR10-DVS.
ImageNet-1K experiments use SDT-8-768, Meta-SpikeFormer-8-512, and
SEW-ResNet18. The semantic segmentation experiment uses an SDT-based
feature pyramid network. Static-image models run for $T=4$ timesteps,
whereas CIFAR10-DVS uses $T=10$.

\subsection{Baselines and Evaluation Protocol}

AdaRound~\cite{nagel2020adaround}, BRECQ~\cite{li2021brecq}, and
GPTQ~\cite{frantar2023gptq} are evaluated as generic weight PTQ
baselines. Their W4/M32 settings quantize weights to 4 bits while
retaining 32-bit membrane states.

For the controlled W4/M4 extension, the W4 weights obtained by each
baseline remain unchanged, and every membrane channel directly reuses
the scale of its preceding weight channel:
\begin{equation}
s_{\mathrm{mem},c}=s_{w,c}.
\end{equation}
No independent membrane observer or additional scale search is
introduced for this Reuse setting. These W4/M4 variants are controlled
extensions implemented in our evaluation and are not configurations
reported in the original baseline papers.

FlowQ~\cite{asim2026flowq} is reimplemented on the same checkpoints
using uniform 4-bit membrane states and layer-wise power-of-two
weight--membrane scale coupling. \method\ instead applies channel-wise
power-of-two coupling and is reported under W4/M4. For \method, M4
denotes an element-count-weighted average membrane precision of
approximately 4 bits; individual membrane channels may use 2, 4, or
8 bits when \mpba\ is enabled.

Within each dataset--backbone pair, all reimplemented methods use the
same pretrained checkpoint, calibration samples, data order, and
first-/last-layer precision. We report Acc@1 for classification and
aAcc, mIoU, and mAcc for semantic segmentation. Accuracy changes are
calculated relative to the corresponding floating-point checkpoint.

\subsection{Quantization and Calibration Settings}

Calibration uses input samples only; ground-truth labels and backbone updates are not used.
Unless otherwise stated, middle convolutional and linear weights use
4 bits. The first projection and classifier remain at higher precision
in the SDT classification experiments.

\subsection{Computing Infrastructure}

Experiments were run with one NVIDIA A800 GPU (80~GB) per run on a
server with Intel Xeon Gold 6336Y CPUs and 1~TB system memory, running
Ubuntu 22.04.5. We used Python 3.10.19, PyTorch 2.7.1 with CUDA 11.8,
TorchVision 0.22.1, timm 0.6.12, and SpikingJelly 0.0.0.0.14.

\section{Additional Classification Results}
\label{app:additional_results}

Table~\ref{tab:appendix_classification} reports supplementary
classification results on CIFAR-10 and CIFAR-100. These experiments
follow the same controlled PTQ protocol as the main-paper comparisons.
A dash denotes a configuration for which a completed evaluation is not
available; no value is extrapolated from another baseline or backbone.

\begin{table*}[t]
\centering
\small
\setlength{\tabcolsep}{7pt}
\begin{tabular}{lllcc}
\toprule
\textbf{Dataset / Backbone} &
\textbf{Method} &
\textbf{M4 scheme} &
\textbf{W4/M32} &
\textbf{W4/M4} \\
\midrule
CIFAR-10 / SDT-2-256
& Full precision
& --
& \multicolumn{2}{c}{\accdrop{94.41}{Base}} \\
\cmidrule(lr){2-5}
& BRECQ~\cite{li2021brecq}
& Reuse
& \accdrop{94.00}{-0.41}
& \accdrop{90.18}{-4.23} \\
& GPTQ~\cite{frantar2023gptq}
& Reuse
& \accdrop{94.07}{-0.34}
& \accdrop{90.93}{-3.48} \\
& FlowQ~\cite{asim2026flowq}
& LW-PoT
& --
& \accdrop{93.57}{-0.84} \\
& \textbf{\method}
& CW-PoT + MPBA
& --
& \textbf{\accdrop{94.35}{-0.06}} \\
\midrule
CIFAR-100 / SDT-2-256
& Full precision
& --
& \multicolumn{2}{c}{\accdrop{76.10}{Base}} \\
\cmidrule(lr){2-5}
& BRECQ~\cite{li2021brecq}
& Reuse
& \accdrop{74.34}{-1.76}
& \accdrop{67.77}{-8.33} \\
& GPTQ~\cite{frantar2023gptq}
& Reuse
& \accdrop{75.90}{-0.20}
& \accdrop{68.82}{-7.28} \\
& FlowQ~\cite{asim2026flowq}
& LW-PoT
& --
& \accdrop{72.59}{-3.51} \\
& \textbf{\method}
& CW-PoT + MPBA
& --
& \textbf{\accdrop{75.28}{-0.82}} \\
\midrule
CIFAR-10 / SEW-ResNet18
& Full precision
& --
& \multicolumn{2}{c}{\accdrop{94.06}{Base}} \\
\cmidrule(lr){2-5}
& BRECQ~\cite{li2021brecq}
& Reuse
& \accdrop{93.45}{-0.61}
& \accdrop{90.71}{-3.35} \\
& GPTQ~\cite{frantar2023gptq}
& Reuse
& \accdrop{93.65}{-0.41}
& \accdrop{90.91}{-3.15} \\
& FlowQ~\cite{asim2026flowq}
& LW-PoT
& --
& \accdrop{91.31}{-2.75} \\
& \textbf{\method}
& CW-PoT + MPBA
& --
& \textbf{\accdrop{93.36}{-0.70}} \\
\bottomrule
\end{tabular}
\normalsize
\caption{Additional classification results. Each accuracy is followed
by its change from the corresponding floating-point checkpoint.
Generic W4/M4 baselines use direct weight-scale reuse, FlowQ uses
layer-wise power-of-two coupling, and \method\ uses channel-wise
coupling under an average 4-bit membrane budget.}
\label{tab:appendix_classification}
\end{table*}

\section{Ablation Studies}
\label{app:ablation}

Unless otherwise stated, all controlled ablations use SEW-ResNet18 on
CIFAR-100 at $T=4$. The floating-point checkpoint reaches 75.27\%
Acc@1, and the first and classifier-adjacent weights use 4 bits.

\subsection{Controlled Ablation Protocol}

\begin{table*}[t]
\centering
\small
\setlength{\tabcolsep}{10pt}
\begin{tabular}{ll@{\qquad}ll}
\toprule
\textbf{Item} &
\textbf{Setting} &
\textbf{Item} &
\textbf{Setting} \\
\midrule
Dataset
& CIFAR-100
& Backbone
& SEW-ResNet18 \\
Timesteps
& $T=4$
& Weight precision
& W4 \\
FP checkpoint
& 75.27
& Calibration samples
& 1024 \\
Scorer seeds
& 0/1/2
& First/head bits
& 4/4 \\
Bridge shift
& integer $k$
& Candidate membrane bits
& 2/4/8 \\
Average membrane budget
& 4.00
& Saturation logging
& enabled \\
\bottomrule
\end{tabular}
\normalsize
\caption{Protocol for the controlled ablation experiments.}
\label{tab:ablation_protocol}
\end{table*}

\section{Resource Accounting}
\label{app:resource_accounting}

The reported membrane-state storage values are theoretical tensor
storage estimates. They do not include allocator fragmentation,
temporary workspace, padding, or mixed-precision packing metadata.

Sequence-level state accounting sums membrane tensors over timesteps,
whereas peak resident memory depends on execution scheduling and buffer
reuse. Consequently, the paper does not claim measured hardware
latency or energy improvements. Such gains require a target-specific
packed integer implementation.

\section{Segmentation Details and Qualitative Results}
\label{app:segmentation}

The Pascal VOC2012 experiment applies the W4/M4 cal\-i\-bra\-tion-only
procedure to an SDT-based FPN without task-specific retraining.
Quantitative segmentation results are reported in the main paper.
This supplementary document provides qualitative comparisons without
duplicating the main result table.

\begin{figure}[ht]
\centering
\includegraphics[width=0.70\columnwidth]{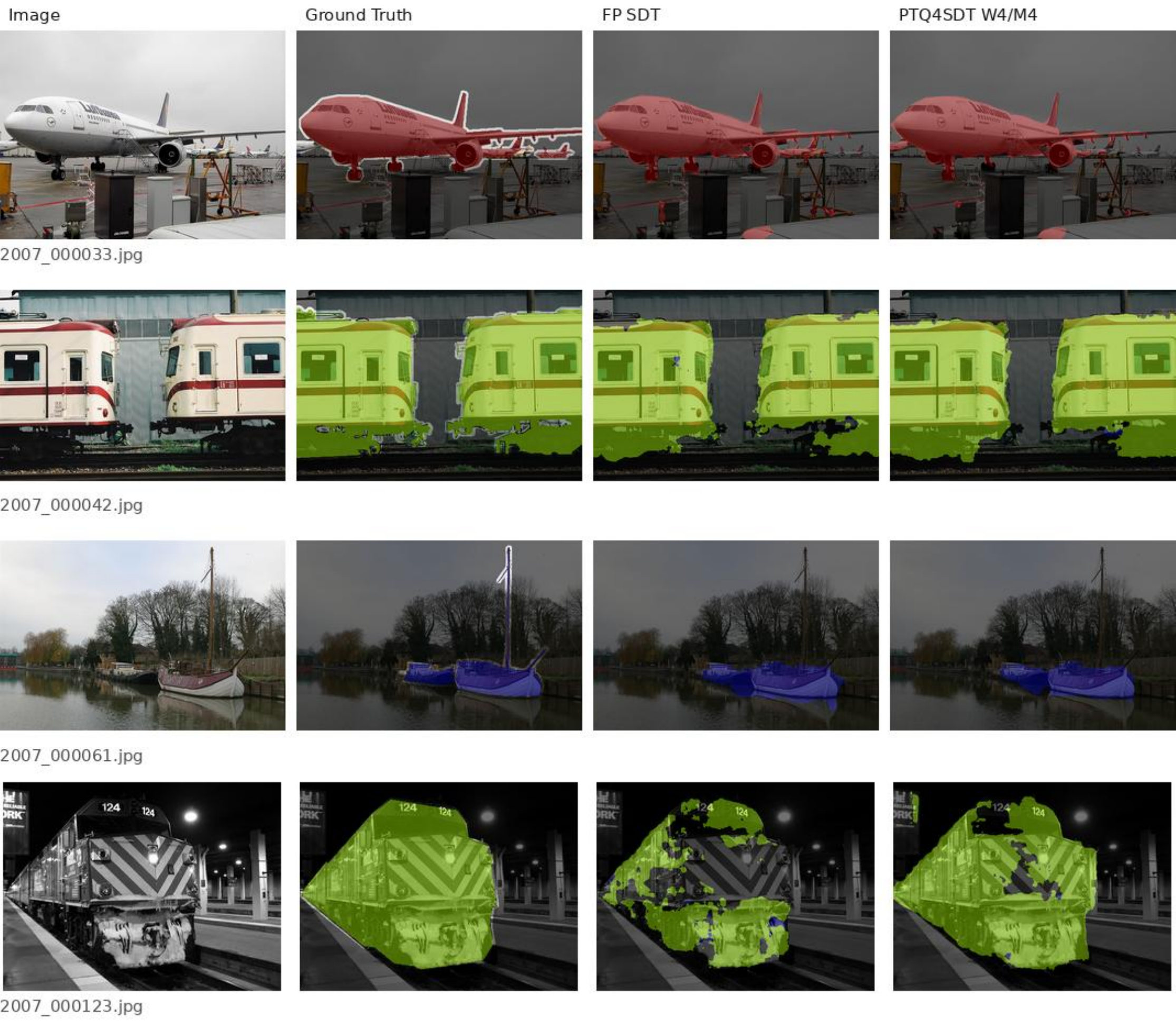}
\caption{Qualitative Pascal VOC2012 examples. Each row shows the input
image, ground truth, floating-point SDT prediction, and \method\ W4/M4
prediction.}
\label{fig:seg_voc_examples}
\end{figure}


\end{document}